\documentclass[11pt,a4paper]{article}
\usepackage{coling2016}
\usepackage{times}
\usepackage{url}
\usepackage{latexsym}

\usepackage[utf8]{inputenc}
\usepackage{times}
\usepackage{url}
\usepackage{latexsym}
\usepackage{color}
\usepackage{tikz}
\usepackage{pgfplots}
\usepackage{lipsum}
\usepackage{units}
\usepackage{amsmath}
\usepackage{amssymb}
\usepackage{relsize}
\usepackage{booktabs}
\usepackage{multirow}
\usepackage{subfigure}
\usepackage[inline]{enumitem}
\usepackage{varwidth}
\usepackage{mathabx}

\usetikzlibrary{arrows, backgrounds, decorations.markings, positioning, shapes}

\newcommand\REVIEW[1]{}
\newcommand\REVIEWOK[1]{}

\title{Keyphrase Annotation with Graph Co-Ranking}

\author{
  Adrien Bougouin \and Florian Boudin \and Béatrice Daille\\
  Université de Nantes, LINA, France\\
  {\normalsize\tt \{adrien.bougouin,florian.boudin,beatrice.daille\}@univ-nantes.fr}\\
}

\date{}

\begin{document}
  \maketitle
  
  \begin{abstract}
    Keyphrase annotation is the task of identifying textual units that represent the main content of a document.
    Keyphrase annotation is either carried out by extracting the most important phrases from a document, keyphrase extraction, or by assigning entries from a controlled domain-specific vocabulary, keyphrase assignment.
    Assignment methods are generally more reliable.
    They provide better-formed keyphrases, as well as keyphrases that do not occur in the document.
    But they are often silent on the contrary of extraction methods that do not depend on manually built resources.
    This paper proposes a new method to perform both keyphrase extraction and keyphrase assignment in an integrated and mutual reinforcing manner.
    Experiments have been carried out on datasets covering different domains of humanities and social sciences.
    They show statistically significant improvements compared to both keyphrase extraction and keyphrase assignment state-of-the art methods.
  \end{abstract}
    
  \section{Introduction}
\label{sec:introduction}

\blfootnote{
    %
    %
    %
    %
    \hspace{-0.65cm}  
    This work is licensed under a Creative Commons 
    Attribution 4.0 International Licence.
    Licence details:
    \url{http://creativecommons.org/licenses/by/4.0/}
    %
    %
}

Keyphrases are words and phrases that give a synoptic picture of what is important within a document.
They are useful in many tasks such as document indexing~\cite{gutwin1999keyphrasesfordigitallibraries}, text categorization~\cite{hulth-megyesi:2006:COLACL} or summarization~\cite{litvak2008graphbased}.
However, most documents do not provide keyphrases, and the daily flow of new documents makes the manual
keyphrase annotation impractical.
As a consequence, automatic keyphrase annotation has received special attention in the NLP community and many methods have been proposed~\cite{hasan2014state_of_the_art}.

The task of automatic keyphrase annotation consists in identifying the main concepts, or topics, addressed in a document.
Such task is crucial to access relevant scientific documents that could be useful for researchers.
Keyphrase annotation methods fall into two broad categories: keyphrase extraction and keyphrase assignment methods.
Keyphrase extraction methods extract the most important words or phrases occurring in a document, while assignment methods provide controlled keyphrases from a domain-specific terminology (controlled vocabulary).

The automatic keyphrase annotation task is often reduced to the sole keyphrase extraction task.
Unlike assignment methods, extraction methods do not require domain specific controlled vocabularies that are costly to create and to maintain.
Furthermore, they are able to identify new concepts that have not been yet recorded in the thesaurus or ontologies.
However, extraction methods often output ill-formed or inappropriate keyphrases~\cite{medelyan2008smalltrainingset}, and they produce only keyphrases that actually occur in the document.

Observations made on manually assigned keyphrases from scientific papers of specialized domains show that professional human indexers both extract keyphrases from the content of the document and assign keyphrases based on their knowledge of the domain~\cite{liu2011vocabularygap}.
Here, we propose an approach that mimics this behaviour and jointly extracts and assigns keyphrases.
We use two graph representations, one for the document and one for the specialized domain.
Then, we apply a co-ranking algorithm to perform both keyphrase extraction and assignment in a mutually reinforcing manner.
%
We perform experiments on bibliographic records in three domains belonging to humanities and social sciences: linguistics, information science and archaeology. 
Along with this approach come two contributions.
First, we present a simple yet efficient assignment extension of a state-of-the-art graph-based keyphrase extraction method, TopicRank~\cite{bougouin2013topicrank}.
Second, we circumvent the need for a controlled vocabulary by leveraging reference keyphrases from training data and further take advantage of their relationship within the training data.

  \section{Related Work}
\label{sec:related_work}
    \subsection{Keyphrase extraction}
    \label{subsec:ake}
        Keyphrase extraction is the most common approach to tackle the automatic keyphrase annotation task. Previous work includes many approaches~\cite{hasan2014state_of_the_art}, from statistical ranking~\cite{salton1975tfidf} to  binary classification~\cite{witten1999kea}, through graph-based ranking~\cite{mihalcea2004textrank} of keyphrase candidates.
        As our approach uses graph-based ranking, we focus on the latter. For a detailed overview of
    keyphrase extraction methods, refer to~\cite{hassan2010conundrums,hasan2014state_of_the_art}.
    
        Since the seminal work of \newcite{mihalcea2004textrank}, graph-based ranking approaches to keyphrase extraction are becoming increasingly popular.
        The original idea behind these approaches is to build a graph from the document and rank its nodes according to their importance using centrality measures.
    
        In TextRank~\cite{mihalcea2004textrank}, the input document is represented as
    a co-occurrence graph in which nodes are words.
    Two words are connected by an edge if they co-occur in a fixed-sized window of words.
    A random walk algorithm is used to iteratively rank the words, then extract the
    keyphrases by concatenating the most important words.

        The random walk algorithm simulates the ``voting concept'', or recommendation: a node is 
    important if it is connected to many other nodes, and if many of those are 
    important.
        Thus, let $G=(V,E)$ be an undirected graph with a set of vertices $V$ and a  set of edges $E$, and let $E(v_i)$ be the set of nodes connected to the node $v_i$.
        The score $S(v_i)$ of a vertex $v_i$ is initialized to 1 and computed iteratively until convergence using the following equation:
        
        \begin{align}
          S(v_i) = (1 - \lambda) + \lambda \sum_{v_j \in E(v_i)}{\frac{S(v_j)}{|E(v_j)|}}
        \end{align}
        
        \noindent where $\lambda$ is a damping factor that has been set to $0.85$ by \newcite{brin1998pagerank} for a trade-off between ranking accuracy and fast convergence.
    
        Following up the work of \newcite{mihalcea2004textrank}, \newcite{wan2008expandrank} added edge weights (co-occurrence numbers) to the random walk and further improved the graph with co-occurrence information borrowed from similar documents. To extract keyphrases from a document, they first look for
    five similar documents, then use them to add new edges between words within
    the graph and reinforce the weight of existing edges.
        \newcite{liu2010topicalpagerank}  biased multiple graphs with topic probabilities drawn from LDA (Latent Dirichlet Allocation)~\cite{blei2003lda}, to rank the words regarding each graph and to merge the rankings together. This method
    performs as many rankings as the number of topics and gives higher importance
    scores to high-ranking words for as many topics as possible. By doing so,
    \newcite{liu2010topicalpagerank} 
    increase the topic coverage provided by the extracted keyphrases.
        %

        Most recently, \newcite{zhang2013wordtopicmultirank} and \newcite{bougouin2013topicrank} explored further the value of topics for keyphrase extraction. 
        \newcite{zhang2013wordtopicmultirank} used graph co-ranking to improve the method of \newcite{liu2010topicalpagerank} by introducing LDA topics right inside the graph.
        \newcite{bougouin2013topicrank} proposed to represent topics as clusters of similar keyphrase candidates within the document (i.e. words and phrases from the document), to rank these topics instead of the words and to extract the most representative candidate as keyphrase for each important topic.
        As our work extends that of \newcite{bougouin2013topicrank}, we present a detailed description of their method in Section~\ref{subsec:topicrank}.
    

    \subsection{Keyphrase assignment}
    \label{subsec:aka}
        Keyphrase assignment provides keyphrases for every document of a specific domain using a controlled vocabulary. Dissimilar to keyphrase extraction, keyphrase assignment also
    aims to provide keyphrases that do not occur within the document. This task is more
    difficult than keyphrase extraction and has, therefore, seldom been employed for automatic
    keyphrase annotation.  The state-of-the art method for keyphrase assignment is KEA++~\cite{medelyan2006kea++}.

        KEA++ uses a domain-specific thesaurus to assign keyphrases to a document.
        First, keyphrase candidates are selected among the n-grams of the document.
        N-grams that do not match a thesaurus entry are either removed or substituted by a synonym that matches a thesaurus entry.
        This candidate selection approach induces a limitation of keyphrase assignment, refered to as keyphrase indexing by \newcite{medelyan2006kea++}, because it only assigns keyphrases if they occur within the document.
        Second, KEA++ exploits the semantic relationships between keyphrase candidates within the thesaurus as the main feature of a Naive Bayes classifier.
        Compared to similar methods without domain specific resources, KEA++ achieves better performance.
        However, such resources are not readily available for most domains, and if so, they could be quickly out of date. 
        The application scenario of KEA++ are thus restricted.
    
  Our proposition is to model  with graphs both keyphrase extraction and assignment and to take benefit of this unified modelling to perform accurate keyphrase annotation. 
  \section{Co-ranking for Keyphrase Annotation}
\label{sec:topicrankpp}
    This section presents TopicCoRank\footnote{TopicCoRank is open source and publicly available at \url{https://github.com/adrien-bougouin/KeyBench/tree/coling_2016/}}, our keyphrase annotation method built on the existing method TopicRank~\cite{bougouin2013topicrank} to which we add keyphrase assignment.
    We first detail TopicRank, then present our contributions.
    
    \subsection{TopicRank}
    \label{subsec:topicrank}
        TopicRank is a graph-based keyphrase extraction method that relies on the following five steps:
        \begin{enumerate}
            \item{\textbf{Keyphrase candidate selection.}
                Following previous work~\cite{hassan2010conundrums,wan2008expandrank}, keyphrase candidates are selected from the sequences of adjacent nouns and adjectives that occur within the document (\texttt{/(N|A)+/}).
            }
            \item{\textbf{Topical clustering.}
                Similar keyphrase candidates $c$ are clustered into topics based on the words they share.
                \newcite{bougouin2013topicrank} use a Hierarchical Agglomerative Clustering (HAC) with a stem overlap similarity (see equation~\ref{math:jaccard}) and an average linkage.
                At the beginning, each keyphrase candidate is a single cluster, then candidates sharing an average of $\unitfrac{1}{4}$ stemmed words with the candidates of another cluster are iteratively added to the latter.
                 
                \vspace{-1em}\begin{align}
                    \text{sim}(c_i, c_j) &= \frac{|\text{stems}(c_i) \cap \text{stems}(c_j)|}{|\text{stems}(c_i) \cup \text{stems}(c_j)|} \label{math:jaccard}
                \end{align}
                where $\text{stems}(c_i)$ is the set of stemmed words of the keyphrase candidate $c_i$.
            }
            \item{\textbf{Graph construction.}
                A complete graph is built, in which nodes are topics and edges are weighted according to the strength of the semantic relation between the connected topics.
                The closer are the pairs of candidates $\langle{}c_i, c_j\rangle{}$ of two topics $t_i$ and $t_j$ within the document, the stronger is their semantic relation $w_{i,j}$:
                
                \vspace{-1em}\begin{align}
                    w_{i, j} &= \mathlarger{\sum}_{c_i \in t_i}\ \mathlarger{\sum}_{c_j \in t_j} \text{dist}(c_i, c_j) \label{math:semantic_relatedness}\\
                    \text{dist}(c_i, c_j) &= \sum_{p_i \in \text{pos}(c_i)}\ \sum_{p_j \in \text{pos}(c_j)} \frac{1}{|p_i - p_j|} \label{math:distance}
                \end{align}
                where $\text{pos}(c_i)$ represents all of the offset positions of the first word of the keyphrase candidate $c_i$.
            }
            \item{\textbf{Topic ranking.}
                Topics $t$ are ranked using the importance score $S(t_i)$ of the Text\-Rank formula, as modified by \newcite{wan2008expandrank} to leverage edge weights:
                 
                \vspace{-1em}\begin{align}
                    S(t_i) = (1 - \lambda) + \lambda \sum_{t_j \in E(t_i)}{\frac{w_{ij}S(t_j)}{\mathlarger\sum_{t_k \in E(t_j)}{w_{jk}}}}\label{math:singlerank}
                \end{align}
            }
            \item{\textbf{Keyphrase selection.}
                One keyphrase candidate is selected from each of the $N$ most important topics: the first occurring keyphrase candidate.
            }
        \end{enumerate}
        
    Our work extends TopicRank to assign domain-specific keyphrases that do not necessarily occur within the document.
    First, we add a second graph representing the domain and unify it to the topic graph.
    Second, we define a co-ranking scheme that leverages the new graph.
    Finally, we redefine the keyphrase selection step for both extracting and assigning keyphrases.

    \subsection{Unified graph construction}
    \label{subsec:graph_construction}
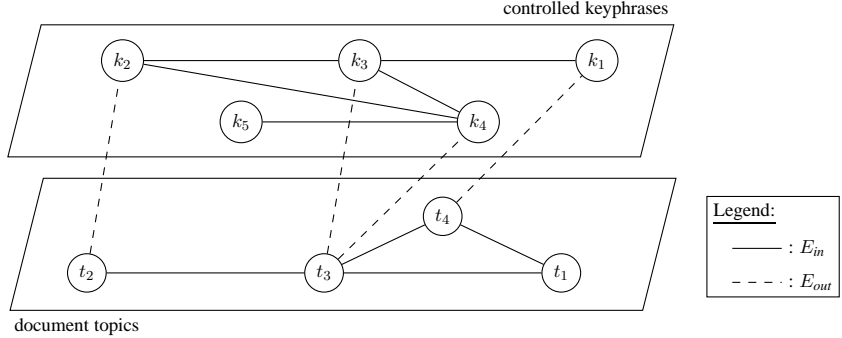
\begin{figure}[htb!]
            \newcommand{\xslant}{0.25}
            \newcommand{\yslant}{0}
        
            \centering
            \begin{tikzpicture}[transform shape, scale=.65]
                \node [draw,
                       rectangle,
                       minimum width=.8\linewidth,
                       minimum height=7em,
                       xslant=\xslant,
                       yslant=\yslant] (domain_graph) {};
                \node [above=of domain_graph,
                       xshift=.4\linewidth,
                       yshift=5.5em,
                       anchor=south east] (domain_graph_label) {controlled keyphrases};
            
                \node [draw,
                       circle,
                       above=of domain_graph,
                       xshift=.3\linewidth,
                       yshift=2.5em] (domain_node1) {$k_1$};
                \node [draw,
                       circle,
                       above=of domain_graph,
                       xshift=-.3\linewidth,
                       yshift=2.5em] (domain_node2) {$k_2$};
                \node [draw,
                       circle,
                       above=of domain_graph,
                       yshift=2.5em] (domain_node3) {$k_3$};
                \node [draw,
                       circle,
                       above=of domain_graph,
                       xshift=.15\linewidth,
                       yshift=-.75em] (domain_node4) {$k_4$};
                \node [draw,
                       circle,
                       above=of domain_graph,
                       xshift=-.15\linewidth,
                       yshift=-.75em] (domain_node5) {$k_5$};
            
                \draw (domain_node1) -- (domain_node3);
                \draw (domain_node2) -- (domain_node3);
                \draw (domain_node2) -- (domain_node4);
                \draw (domain_node4) -- (domain_node5);
                \draw (domain_node4) -- (domain_node3);
            
                \node [draw,
                       rectangle,
                       minimum width=.8\linewidth,
                       minimum height=7em,
                       xslant=\xslant,
                       yslant=\yslant,
                       above=of domain_graph,
                       xshift=-1.5em,
                       yshift=-1.5em] (document_graph) {};
                \node [below=of document_graph,
                       xshift=-.4\linewidth,
                       yshift=-5.56em,
                       anchor=north west] (document_graph_label) {document topics};
            
                \node [draw,
                       circle,
                       below=of document_graph,
                       xshift=.3\linewidth,
                       yshift=-2.5em] (document_node1) {$t_1$};
                \node [draw,
                       circle,
                       below=of document_graph,
                       xshift=-.3\linewidth,
                       yshift=-2.5em] (document_node2) {$t_2$};
                \node [draw,
                       circle,
                       below=of document_graph,
                       yshift=-2.5em] (document_node3) {$t_3$};
                \node [draw,
                       circle,
                       below=of document_graph,
                       xshift=.15\linewidth,
                       yshift=.5em] (document_node4) {$t_4$};
            
                \draw (document_node2) -- (document_node3);
                \draw (document_node3) -- (document_node1);
                \draw (document_node1) -- (document_node4);
                \draw (document_node3) -- (document_node4);
            
                \draw [dashed] (document_node2) -- (domain_node2);
                \draw [dashed] (document_node3) -- (domain_node3);
                \draw [dashed] (document_node4) -- (domain_node1);
                \draw [dashed] (document_node3) -- (domain_node4);
            
                \node [right=of document_graph, yshift=-6.5em] (legend_title) {\underline{Legend:}};
                \node [below=of legend_title, xshift=-1em, yshift=2em] (begin_inner) {};
                \node [right=of begin_inner] (end_inner) {: $E_\textnormal{\textit{in}}$};
                \node [below=of begin_inner, yshift=1.5em] (begin_outer) {};
                \node [right=of begin_outer] (end_outer) {: $E_\textnormal{\textit{out}}$};
            
                \draw (legend_title.north  -| end_outer.east) rectangle (end_outer.south -| legend_title.west);
            
                \draw (begin_inner) -- (end_inner);
                \draw [dashed] (begin_outer) -- (end_outer);
            \end{tikzpicture}
            \caption{
                Example of a unified graph constructed by TopicCoRank and its two kinds of edges
                \label{fig:topicrankpp_graph}
            }
        \end{figure}

        TopicCoRank operates over a unified graph that connects two graphs representing the document topics, the controlled keyphrases and the relations between them (see Fig.~\ref{fig:topicrankpp_graph}).
        The controlled keyphrases are the keyphrases that were manually assigned to training documents.
        Considering the manually assigned keyphrases as the controlled vocabulary circumvents the need for a manually produced controlled vocabulary and also allows us to further take advantage of the semantic relatonship between the domain-specific (controlled) keyphrases.
        Because controlled keyphrases are presumably non-redundant, we do not topically cluster them as we do for keyphrase candidates.
        
        \REVIEWOK{The notations of the vertices for the controlled keyphrases and topics are very confused.}
        Let $G = (V=T \cup K, E=E_{\textnormal{\textit{in}}} \cup E_{\textnormal{\textit{out}}})$ denote the unified graph.
        Topics $T=\{t_1, t_2, ..., t_n\}$ and controlled keyphrases $K=\{k_1, k_2, ..., k_m\}$ are vertices $V$ connected to their fellows by edges $E_\textnormal{\textit{in}} \subseteq T \times T \cup K \times K$ and connected to the other vertices by edges $E_\textnormal{\textit{out}} \subseteq K \times T$ (see Fig.~\ref{fig:topicrankpp_graph}).

        To unify the two graphs, we consider the controlled keyphrases as a category map and connect the document to its potential categories.
        We create an unweighted edge $\langle{}k_i, t_j\rangle{} \in E_{\textnormal{\textit{out}}}$ to connect a controlled keyphrase $k_i$ and a topic $t_j$ if the controlled keyphrase is a member of the topic, i.e. a keyphrase candidate of the topic\footnote{To accept inflexions, such as plural inflexions, we follow \newcite{bougouin2013topicrank} and perform the comparison with stems.}.
        We create an edge $\langle{}t_i, t_j\rangle{} \in E_\textnormal{\textit{in}}$ or $\langle{}k_i, k_j\rangle{} \in E_\textnormal{\textit{in}}$ between two topics $t_i$ and $t_j$ or two controlled keyphrases $k_i$ and $k_j$ when they co-occur within a sentence of the document or as keyphrases of a training document, respectively.
        Edges $\langle{}t_i, t_j\rangle{} \in E_{\textnormal{\textit{in}}}$ are weighted by the number of times ($w_{i, j}$) topics $t_i$ and $t_j$ occur in the same sentence within the document.
        Edges $\langle{}k_i, k_j\rangle{} \in E_{\textnormal{\textit{in}}}$ are weighted by the number of times ($w_{i, j}$) keyphrases $k_i$ and $k_j$ are associated to the same document among the training documents.
        Doing so, the weighting scheme of edges $E_\textnormal{\textit{in}}$ is equivalent for both topics and controlled keyphrases.
        This equivalence is essential to ensure that not only controlled keyphrases occurring in the document can be assigned by properly co-ranking topics and controlled keyphrases.

    \subsection{Graph-based co-ranking}
    \label{subsec:graph_based_co_ranking}
        TopicCoRank gives an importance score $S(t_i)$ or $S(k_i)$ to every topic or controlled keyphrase using graph co-ranking (see equations~\ref{math:topiccorank_t} and~\ref{math:topiccorank_k}).
        %
        Our graph co-ranking simulates the voting concept based on inner and outer recommendations.
        
        The inner recommendation is similar to the recommendation computed in previous work~\cite{bougouin2013topicrank,mihalcea2004textrank,wan2008expandrank}.
        The inner recommendation $R_\textnormal{\textit{in}}$ comes from nodes of the same graph (see equation~\ref{math:rin}). 
        A topic or a controlled keyphrase is important if it is strongly connected to other topics or controlled keyphrases, respectively.
        
        The outer recommendation influences the ranking of topics by controlled keyphrases and of controlled keyphrases by topics.
        The outer recommendation $R_\textnormal{\textit{out}}$ comes from nodes of the other graph (see equation~\ref{math:rout}). 
        A topic or a controlled keyphrase gain more importance if it is connected to important controlled keyphrases or an important topic, respectively.
        
        \begin{align}
          S(t_i) &= (1 - \lambda_t)\ R_{out}(t_i) + \lambda_t\ R_{in}(t_i)\label{math:topiccorank_t}\\[1em]
          S(k_i) &= (1 - \lambda_k)\ R_{out}(k_i) + \lambda_k\ R_{in}(k_i)\label{math:topiccorank_k}\\[1em]
          R_{in}(v_i) &= \sum_{v_j \in E_{\text{in}}(v_i)}{\frac{w_{ij} S(v_j)}{\mathlarger\sum_{v_k \in E_{\text{in}}(v_j)}{{w_{jk}}}}}\label{math:rin}\\[1em]
          R_{out}(v_i) &= \sum_{v_j \in E_{\text{out}}(v_i)}{\frac{S(v_j)}{|E_{\text{\textit{out}}}(v_j)|}}\label{math:rout}
        \end{align}

        \noindent where $v_i$ is a node representing a keyphrase or a topic.
        $\lambda_t$ and $\lambda_k$ are parameters that control the influence of the inner recommendation over the outer recommendation ($0~\leq~\lambda_t~\leq~1$ and $0~\leq~\lambda_k~\leq~1$) for the topics and the controlled keyphrases, respectively.

    \subsection{Keyphrase annotation}
    \label{subsec:keyphrase_assignment_and_extraction}
        Keyphrases are extracted and assigned from the N-best ranked topics and controlled keyphrases, regardless of their nature.
        
        We extract topic keyphrases using the former TopicRank strategy. Only one keyphrase is extracted per topic: the keyphrase candidate that first occurs within the document.
        
        We assign controlled keyphrases only if they are directly or transitively connected to a topic of the document. If the ranking of a controlled keyphrase has not been affected by a topic of the document nor by controlled keyphrases connected to topics, then its importance score is not related to the content of the document and it should not be assigned.

        At this step, two variants of TopicCoRank performing either extraction or assignment can be proposed, namely TopicCoRank$_\textit{extr}$ and TopicCoRank$_\textit{assign}$.
        If keyphrases are only extracted from the topics, we obtain TopicCoRank$_\textit{extr}$.
        If keyphrases are only assigned from the controlled keyphrases, we obtain TopicCoRank$_\textit{assign}$.

  \section{Experimental Setup}
\label{sec:experimental_setup}
    \subsection{Datasets}
    \label{subsec:datasets}
    
        We conduct our experiments on data from the DEFT-2016 benchmark datasets~\cite{daille-et-al:2016:DEFT}\footnote{Data has been provided by the TermITH project for both DEFT-2016 and this work. Parallely, the subset division has been modified for the purpose of DEFT-2016. Therefore, we use the same data as DEFT-2016, but the subset division is different. The subset division we used for our experiences can be found here: \url{https://github.com/adrien-bougouin/KeyBench/tree/coling_2016/datasets/}} in three domains: linguistics, information Science and archaeology.
        Table~\ref{tab:dataset_statistics} shows the factual information about the datasets.
        Each dataset is a collection of 706 up to 718 French bibliographic records collected from the database of the French Institute for Scientific and Technical Information\footnote{\url{http://www.inist.fr}} (Inist).
        The bibliographic records contain a title of one scientific paper, its abstract and its keyphrases that were annotated by professional indexers (one per bibliographic record).
        Indexers were given the instruction to assign reference keyphrases from a controlled vocabulary and to extract new concepts or very specific keyphrases from the titles and the abstracts.
        Each dataset is divided into three sets: a test set, used for evaluation; a training set (denoted as train), used to represent the domain; and a development set (denoted as dev), used for parameter tuning.
        %
        
            \begin{table}[htb!]
            \centering
            \begin{tabular}{l|c@{\hspace{1em}}c@{\hspace{1em}}c@{\hspace{.5em}}|@{\hspace{.5em}}c@{\hspace{1em}}c@{\hspace{1em}}c@{\hspace{.5em}}|@{\hspace{.5em}}c@{\hspace{1em}}c@{\hspace{1em}}c}
            \toprule
                \multirow{2}{*}{\textbf{Corpus}} & \multicolumn{3}{c@{\hspace{.5em}}|@{\hspace{.5em}}}{\textbf{Linguistics}} & \multicolumn{3}{c@{\hspace{.5em}}|@{\hspace{.5em}}}{\textbf{Information Science}} & \multicolumn{3}{c}{\textbf{Archaeology}}\\
                \cline{2-10}
                & \multicolumn{1}{c@{~$\supset$}}{train} & dev & test & \multicolumn{1}{c@{~$\supset$}}{train} & dev & test & \multicolumn{1}{c@{~$\supset$}}{train} & dev & test\\
                \hline
                Documents & 515 & 100 & 200  & 506 & 100 & 200 & 518 & 100 & 200\\
                Tokens/Document & 161 & 151 & 147 & 105 & 152 & 157 & 221 & 201 & 214\\
                Keyphrases & 8.6 & 8.8 & 8.9 & 7.8 & 10.0 & 10.2 & 16.9 & 16.4 & 15.6\\
                Missing Keyphrases (\%) & 60.6 & 63.2 & 62.8 & 67.9 & 63.1 & 66.9 & 37.0 & 48.4 & 37.4\\
                \bottomrule
            \end{tabular}
            \caption{
                Dataset statistics.
                ``Missing'' represents the percentage of keyphrases that cannot be retrieved within the documents.
                \label{tab:dataset_statistics}}
        \end{table}

        The amount of missing keyphrases, i.e.~keyphrases that cannot be extracted from the documents, shows the importance of keyphrase assignment in the context of scientific domains.
        More than half of the keyphrases of linguistics and information science domains can only be assigned, which confirms that these two datasets are difficult to process with keyword extraction approaches alone. 

    \subsection{Document preprocessing}
    \label{subsec:document_preprocessing}
        We apply the following preprocessing steps to each document: sentence segmentation, word tokenization and Part-of-Speech (POS) tagging.
        Sentence segmentation is performed with the PunktSentenceTokenizer provided by the Python Natural Language ToolKit (NLTK)~\cite{bird2009nltk}, word tokenization using the Bonsai word tokenizer\footnote{The Bonsai word tokenizer is a tool provided with the Bonsai PCFG-LA parser: \url{http://alpage.inria.fr/statgram/frdep/fr_stat_dep_parsing.html}} and POS tagging with MElt~\cite{denis2009melt}.

    \subsection{Baselines}
    \label{subsec:baselines}
        To show the effectiveness of our approach, we compare TopicCoRank and its variants (TopicCoRank$_\textnormal{\textit{extr}}$ and TopicCoRank$_\textnormal{\textit{assign}}$) with TopicRank and KEA++.
        For KEA++, we use the thesauri maintained by Inist\footnote{Thesauri are available from: \url{http://deft2016.univ-nantes.fr/download/traindev/}} to index the bibliographic records of Linguistics, Information Science and Archaeology.
    
    

    \subsection{TopicCoRank setting}
    \label{subsec:topiccorank_settings}
        \REVIEWOK{Empirically, we'd tune the parameters on development set, which is independant of evaluation dataset}
        The $\lambda_t$ and $\lambda_k$ parameters of TopicCoRank were tuned on the development sets, and set to 0.1 and 0.5 respectively.
        This empirical setup means that the importance of topics is much more influenced by controlled keyphrases than other topics, and that the importance of controlled keyphrases is equally influenced by controlled keyphrases and topics.
        In other words, the domain has a positive influence on the joint task of keyphrase extraction and assignment.

  \section{Experimental Results}
\label{sec:experimental_results}
    This section presents and analyses the results of our experiments.
    For each document of each dataset, we compare the keyphrases outputed by each method to the reference keyphrases of the document.
    From the comparisons, we compute the macro-averaged precision (P), recall (R) and f1-score (F) per dataset and per method.
        
    \subsection{Macro-averages results}
    \label{subsec:macro_averages_results}
        Table~\ref{tab:comparison_results} presents the macro-averaged precision, recall and f1-score in percentage when 10 keyphrases are extracted/assigned for each dataset by TopicRank, KEA++, TopicCo\-Rank$_{\textit{extr}}$, Topic\-CoRank$_{\textit{assign}}$ and TopicCoRank.
        First, we observe that the assignment baseline KEA++ mostly achieves the lowest performance, which is surprising compared to the performance reported by \newcite{medelyan2006kea++}.
        The first reason for this observation is that KEA++ is restricted to thesauri entries while most keyphrases are missing within our documents.
        The second reason is that KEA++ relies on rich thesauri that contain an important amount of semantic relations between the entries, while our (real application) thesauri have a modest amount of semantic relations between the entries.

        Overall, using graph co-ranking significantly outperforms TopicRank and KEA++.
        Comparing TopicRank to TopicCoRank$_\textit{extr}$ shows the positive influence of the domain (controlled keyphrases) on the ranking of the topics.
        TopicCoRank$_\textit{assign}$ outperforms every method, including TopicCoRank$_\textit{extr}$ and TopicCoRank.
        Controlled keyphrases are efficiently ranked and the predominance of missing keyphrases in the dataset leads to a better performance of TopicCoRank$_\textit{assign}$ over TopicCoRank.
        
        \begin{table}[htb!]
            \centering
            \begin{tabular}{l|c@{\hspace{1em}}c@{\hspace{1em}}c@{\hspace{.5em}}|@{\hspace{.5em}}c@{\hspace{1em}}c@{\hspace{1em}}c@{\hspace{.5em}}|@{\hspace{.5em}}c@{\hspace{1em}}c@{\hspace{1em}}c}
                \toprule
                \multirow{2}{*}{\textbf{Method}} & \multicolumn{3}{c@{\hspace{.5em}}|@{\hspace{.5em}}}{\textbf{Linguistics}} & \multicolumn{3}{c@{\hspace{.5em}}|@{\hspace{.5em}}}{\textbf{Information Science}} & \multicolumn{3}{c}{\textbf{Archaeology}}\\
                \cline{2-10}
                & P & R & F$^{~~}$ & P & R & F$^{~~}$ & P & R & F\\
                \hline
                TopicRank & 11.82 & 13.1 & 11.9$^{~~}$ & 12.1 & 12.8 & 12.1 & 27.5 & 19.7 & 21.8$^{~~}$\\
                KEA++ & 11.6 & 13.0 & 12.1$^{~~}$ & $~~$9.5 & 10.2 & $~~$9.6 & 23.5 & 16.2 & 18.8$^{~~}$\\
                \hline
                TopicCoRank$_\textnormal{\textit{extr}}$ & 15.9 & 18.2 & 16.7$^{\dagger}$ & 15.9 & 16.2 & 15.6$^{\dagger}$ & 39.6 & 26.4 & 31.0$^{\dagger}$\\
                TopicCoRank$_\textnormal{\textit{assign}}$ & \textbf{25.8} & \textbf{29.6} & \textbf{27.2}$^{\dagger}$ & \textbf{19.9} & \textbf{20.0} & \textbf{19.5}$^{\dagger}$ & \textbf{49.6} & \textbf{33.3} & \textbf{39.0}$^{\dagger}$\\
                \hline
                TopicCoRank & 24.5 & 28.3 & 25.9$^{\dagger}$ & 19.4 & 19.6 & 19.0$^{\dagger}$ & 46.6 & 31.4 & 36.7$^{\dagger}$\\
                \bottomrule
            \end{tabular}
            \caption{
                Results of TopicCoRank and the baselines at 10 keyphrases for each dataset.
                Precision (P), Recall (R) and F-score (F) are reported in percentages. $\dagger$ indicates a significant F-score improvement over TopicRank and KEA++ at 0.001 level using Student’s t-test.
                \label{tab:comparison_results}}
        \end{table}
  
    \subsection{Precision/recall curves}
    \label{subsec:precision_recall_curves}
        Additionally, we follow \newcite{hassan2010conundrums} and analyse the precision-recall curves of TopicRank, KEA++ and TopicCoRank.
        To generate the curves, we vary the number of evaluated keyphrases (cut-off) from 1 to the total number of extracted/assigned key\-phrases and compute the precision and recall for each cut-off.
        Such representation gives a good appreciation of the advantage of a method compared to others, especially if the other methods achieve performances in the \textit{Area Under the Curve} (AUC).
        
         \begin{figure}[htb!]
            \centering
            \subfigure[Linguistics]{
                \begin{tikzpicture}[scale=.7]
                    \pgfkeys{/pgf/number format/.cd, fixed}
                    \begin{axis}[x=0.0050692\linewidth,
                                 xtick={0, 10, 20, ..., 100},
                                 xmin=0,
                                 xmax=60,
                                 xlabel=recall (\%),
                                 x label style={yshift=.34em},
                                 y=0.0050692\linewidth,
                                 ytick={0, 10, 20, ..., 100},
                                 ymin=0,
                                 ymax=60,
                                 ylabel=precision (\%),
                                 y label style={yshift=-.5em}]
                        \addplot [cyan, mark=+] file {input/data/linguistique_topicrank.csv};
                        \addplot [magenta, mark=o] file {input/data/linguistique_kea_pp.csv};
                        \addplot [blue, mark=x] file {input/data/linguistique_topiccorank.csv};
                        \addplot [dotted, domain=30:60] {(50 * x) / ((2 * x) - 50)};
                        \addplot [dotted, domain=30:60] {(40 * x) / ((2 * x) - 40)};
                        \addplot [dotted, domain=20:60] {(30 * x) / ((2 * x) - 30)};
                        \addplot [dotted, domain=10:60] {(20 * x) / ((2 * x) - 20)};
                        \addplot [dotted, domain=5:60] {(10 * x) / ((2 * x) - 10)};
                        \legend{TopicRank, KEA++, TopicCoRank};
                    \end{axis}
                    \node at (4.9,2.6) [anchor=east] {\scriptsize{F=40.0}};
                    \node at (4.9,1.8) [anchor=east] {\scriptsize{F=30.0}};
                    \node at (4.9,1.15) [anchor=east] {\scriptsize{F=20.0}};
                    \node at (4.9,0.6) [anchor=east] {\scriptsize{F=10.0}};
                \end{tikzpicture}
            }
            \subfigure[Information Science]{
                \begin{tikzpicture}[scale=.7]
                    \pgfkeys{/pgf/number format/.cd, fixed}
                    \begin{axis}[x=0.0050692\linewidth,
                                 xtick={0, 10, 20, ..., 100},
                                 xmin=0,
                                 xmax=60,
                                 xlabel=recall (\%),
                                 x label style={yshift=.34em},
                                 y=0.0050692\linewidth,
                                 ytick={0, 10, 20, ..., 100},
                                 ymin=0,
                                 ymax=60,
                                 ylabel=precision (\%),
                                 y label style={yshift=-.5em}]
                        \addplot [cyan, mark=+] file {input/data/sciences_de_l_information_topicrank.csv};
                        \addplot [magenta, mark=o] file {input/data/sciences_de_l_information_kea_pp.csv};
                        \addplot [blue, mark=x] file {input/data/sciences_de_l_information_topiccorank.csv};
                        \addplot [dotted, domain=30:60] {(50 * x) / ((2 * x) - 50)};
                        \addplot [dotted, domain=30:60] {(40 * x) / ((2 * x) - 40)};
                        \addplot [dotted, domain=20:60] {(30 * x) / ((2 * x) - 30)};
                        \addplot [dotted, domain=10:60] {(20 * x) / ((2 * x) - 20)};
                        \addplot [dotted, domain=5:60] {(10 * x) / ((2 * x) - 10)};
                    \end{axis}
                    \node at (4.9,3.7) [anchor=east] {\scriptsize{F=50.0}};
                    \node at (4.9,2.6) [anchor=east] {\scriptsize{F=40.0}};
                    \node at (4.9,1.8) [anchor=east] {\scriptsize{F=30.0}};
                    \node at (4.9,1.15) [anchor=east] {\scriptsize{F=20.0}};
                    \node at (4.9,0.6) [anchor=east] {\scriptsize{F=10.0}};
                \end{tikzpicture}
            }
            \subfigure[Archaeology]{
                \begin{tikzpicture}[scale=.7]
                    \pgfkeys{/pgf/number format/.cd, fixed}
                    \begin{axis}[x=0.0050692\linewidth,
                                 xtick={0, 10, 20, ..., 100},
                                 xmin=0,
                                 xmax=60,
                                 xlabel=recall (\%),
                                 x label style={yshift=.34em},
                                 y=0.0050692\linewidth,
                                 ytick={0, 10, 20, ..., 100},
                                 ymin=0,
                                 ymax=60,
                                 ylabel=precision (\%),
                                 y label style={yshift=-.5em},
                                 legend style={font=\footnotesize}]
                        \addplot [cyan, mark=+] file {input/data/archeologie_topicrank.csv};
                        \addplot [magenta, mark=o] file {input/data/archeologie_kea_pp.csv};
                        \addplot [blue, mark=x] file {input/data/archeologie_topiccorank.csv};
                        \addplot [dotted, domain=30:60] {(50 * x) / ((2 * x) - 50)};
                        \addplot [dotted, domain=30:60] {(40 * x) / ((2 * x) - 40)};
                        \addplot [dotted, domain=20:60] {(30 * x) / ((2 * x) - 30)};
                        \addplot [dotted, domain=10:60] {(20 * x) / ((2 * x) - 20)};
                        \addplot [dotted, domain=5:60] {(10 * x) / ((2 * x) - 10)};
                    \end{axis}
                    \node at (4.9,3.7) [anchor=east] {\scriptsize{F=50.0}};
                    \node at (4.9,2.6) [anchor=east] {\scriptsize{F=40.0}};
                    \node at (4.9,1.8) [anchor=east] {\scriptsize{F=30.0}};
                    \node at (4.9,1.15) [anchor=east] {\scriptsize{F=20.0}};
                    \node at (4.9,0.6) [anchor=east] {\scriptsize{F=10.0}};
                \end{tikzpicture}
            }
            \caption{
                Precision-recall curves of TopicRank, KEA++ and TopicCoRank for each dataset
                \label{fig:pr_curves}
            }
        \end{figure}
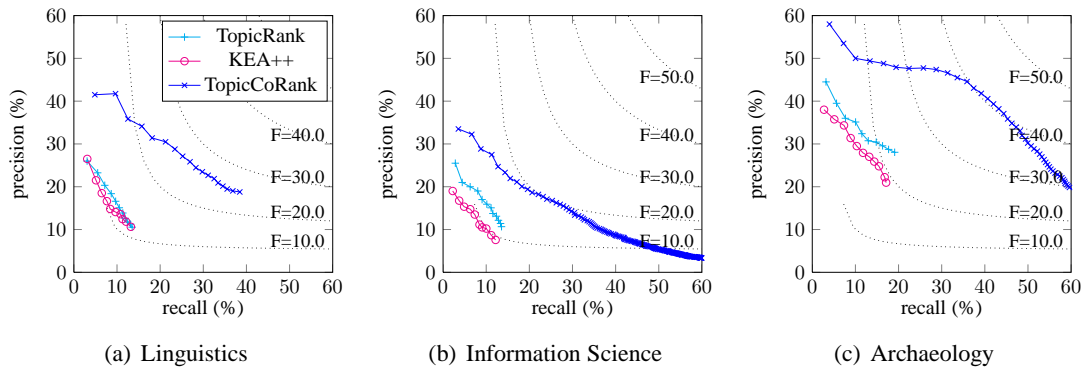

        Figure~\ref{fig:pr_curves} shows the precision/recall curves of TopicRank, KEA++ and TopicCoRank on each dataset.
        The final recall for the methods does not reach 100\% because the candidate selection method does not provide keyphrases that do not occur within the document, as well as candidates that do not fit the POS tag pattern \texttt{/(N|A)+/}.
        Also, because TopicRank and TopicCoRank topically cluster keyphrase candidates and output only one candidate per topic, their final recall is lowered every time a wrong keyphrase is chosen over a correct one from the topic.
 
        We observe that the curve for TopicCoRank is systematically above the others, thus showing improvements in the area under the curve and not just in point estimate such as f1-score.
        Also, the final recall of TopicCoRank is much higher than the final recall of TopicRank and KEA++.

    \subsection{Extraction vs. assignment}
    \label{subsec:assignment_vs_extraction}
        As TopicCoRank is the first method for simultaneously extracting and assigning key\-phrases, we perform an additional experiment that shows to which extent extraction and assignment contribute to the final results.
        %
        To do so, we show the behavior of the extraction and the assignment depending on the influence of the inner recommendation on the ranking for each (test) document of each dataset.
        
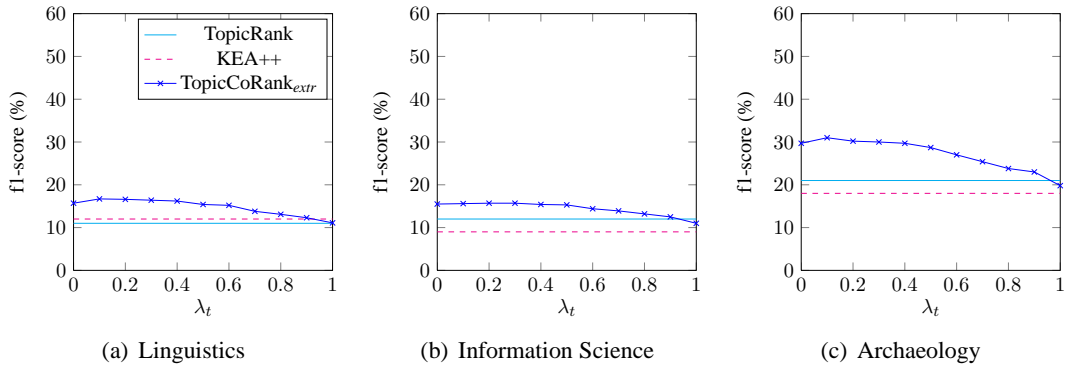
\begin{figure}[htb!]
            \centering
            \subfigure[Linguistics]{
                \begin{tikzpicture}[scale=.7]
                    \pgfkeys{/pgf/number format/.cd, fixed}
                    \begin{axis}[x=0.304152\linewidth,
                                 xtick={0, 0.2, 0.4, ..., 1},
                                 xmin=0,
                                 xmax=1,
                                 xlabel=$\lambda_t$,
                                 x label style={yshift=.34em},
                                 y=0.0050692\linewidth,
                                 ytick={0, 10, 20, ..., 100},
                                 ymin=0,
                                 ymax=60,
                                 ylabel=f1-score (\%),
                                 y label style={yshift=-.5em}]
                        \addplot[cyan] coordinates{
                            (0.0, 11)
                            (0.2, 11)
                            (0.3, 11)
                            (0.4, 11)
                            (0.5, 11)
                            (0.6, 11)
                            (0.7, 11)
                            (0.8, 11)
                            (0.9, 11)
                            (1.0, 11)
                        };
                        \addplot[magenta, dashed] coordinates{
                            (0.0, 12)
                            (0.2, 12)
                            (0.3, 12)
                            (0.4, 12)
                            (0.5, 12)
                            (0.6, 12)
                            (0.7, 12)
                            (0.8, 12)
                            (0.9, 12)
                            (1.0, 12)
                        };
                        \addplot[blue, mark=x] coordinates{
                            (0.0, 15.7)
                            (0.1, 16.7)
                            (0.2, 16.6)
                            (0.3, 16.4)
                            (0.4, 16.2)
                            (0.5, 15.4)
                            (0.6, 15.2)
                            (0.7, 13.8)
                            (0.8, 13.1)
                            (0.9, 12.3)
                            (1.0, 11.1)
                        };
                        \legend{TopicRank, KEA++, TopicCoRank$_\textit{extr}$};
                    \end{axis}
                \end{tikzpicture}
            }
            \subfigure[Information Science]{
                \begin{tikzpicture}[scale=.7]
                    \pgfkeys{/pgf/number format/.cd, fixed}
                    \begin{axis}[x=0.304152\linewidth,
                                 xtick={0, 0.2, 0.4, ..., 1},
                                 xmin=0,
                                 xmax=1,
                                 xlabel=$\lambda_t$,
                                 x label style={yshift=.34em},
                                 y=0.0050692\linewidth,
                                 ytick={0, 10, 20, ..., 100},
                                 ymin=0,
                                 ymax=60,
                                 ylabel=f1-score (\%),
                                 y label style={yshift=-.5em}]
                        \addplot[cyan] coordinates{
                            (0.0, 12)
                            (0.2, 12)
                            (0.3, 12)
                            (0.4, 12)
                            (0.5, 12)
                            (0.6, 12)
                            (0.7, 12)
                            (0.8, 12)
                            (0.9, 12)
                            (1.0, 12)
                        };
                        \addplot[magenta, dashed] coordinates{
                            (0.0, 9)
                            (0.2, 9)
                            (0.3, 9)
                            (0.4, 9)
                            (0.5, 9)
                            (0.6, 9)
                            (0.7, 9)
                            (0.8, 9)
                            (0.9, 9)
                            (1.0, 9)
                        };
                        \addplot[blue, mark=x] coordinates{
                            (0.0, 15.5)
                            (0.1, 15.6)
                            (0.2, 15.7)
                            (0.3, 15.7)
                            (0.4, 15.4)
                            (0.5, 15.3)
                            (0.6, 14.4)
                            (0.7, 13.9)
                            (0.8, 13.2)
                            (0.9, 12.5)
                            (1.0, 11.0)
                        };
                    \end{axis}
                \end{tikzpicture}
            }
            \subfigure[Archaeology]{
                \begin{tikzpicture}[scale=.7]
                    \pgfkeys{/pgf/number format/.cd, fixed}
                    \begin{axis}[x=0.304152\linewidth,
                                 xtick={0, 0.2, 0.4, ..., 1},
                                 xmin=0,
                                 xmax=1,
                                 xlabel=$\lambda_t$,
                                 x label style={yshift=.34em},
                                 y=0.0050692\linewidth,
                                 ytick={0, 10, 20, ..., 100},
                                 ymin=0,
                                 ymax=60,
                                 ylabel=f1-score (\%),
                                 y label style={yshift=-.5em},
                                 legend style={font=\footnotesize}]
                        \addplot[cyan] coordinates{
                            (0.0, 21)
                            (0.2, 21)
                            (0.3, 21)
                            (0.4, 21)
                            (0.5, 21)
                            (0.6, 21)
                            (0.7, 21)
                            (0.8, 21)
                            (0.9, 21)
                            (1.0, 21)
                        };
                        \addplot[magenta, dashed] coordinates{
                            (0.0, 18)
                            (0.2, 18)
                            (0.3, 18)
                            (0.4, 18)
                            (0.5, 18)
                            (0.6, 18)
                            (0.7, 18)
                            (0.8, 18)
                            (0.9, 18)
                            (1.0, 18)
                        };
                        \addplot[blue, mark=x] coordinates{
                            (0.0, 29.7)
                            (0.1, 31.0)
                            (0.2, 30.2)
                            (0.3, 30.0)
                            (0.4, 29.7)
                            (0.5, 28.7)
                            (0.6, 27.0)
                            (0.7, 25.4)
                            (0.8, 23.8)
                            (0.9, 23.0)
                            (1.0, 19.8)
                        };
                    \end{axis}
                \end{tikzpicture}
            }
            \caption{
                Behavior of TopicCoRank$_\textit{extr}$ depending on $\lambda_t$ ($\lambda_k = 0.5$)
                \label{fig:lambda_t_variation}
            }
        \end{figure}

        Fig.~\ref{fig:lambda_t_variation} shows the behavior of TopicCoRank$_\textit{extr}$ when $\lambda_t$ varies from 0 to 1.
        When $\lambda_t = 0$, only the domain influences the ranking of the topics.
        Slightly equivalent to KEA++, TopicCoRank$_\textit{extr}$ with $\lambda_t = 0$ mainly extracts keyphrases from topics connected to controlled keyphrases.
        When $\lambda_t = 1$, the domain does not influence the ranking and the performance of TopicCoRank$_\textit{extr}$ is in the range of TopicRank's performance.
        Overall, the performance curve of TopicCoRank$_\textit{extr}$ decreases while $\lambda_t$ increases.
        Thus, the experiment demonstrates that the domain has a positive influence on the keyphrase extraction.
        
\begin{figure}[htb!]
            \centering
            \subfigure[Linguistics]{
                \begin{tikzpicture}[scale=.7]
                    \pgfkeys{/pgf/number format/.cd, fixed}
                    \begin{axis}[x=0.304152\linewidth,
                                 xtick={0, 0.2, 0.4, ..., 1},
                                 xmin=0,
                                 xmax=1,
                                 xlabel=$\lambda_k$,
                                 x label style={yshift=.34em},
                                 y=0.0050692\linewidth,
                                 ytick={0, 10, 20, ..., 100},
                                 ymin=0,
                                 ymax=60,
                                 ylabel=f1-score (\%),
                                 y label style={yshift=-.5em}]
                        \addplot[cyan] coordinates{
                            (0.0, 11)
                            (0.2, 11)
                            (0.3, 11)
                            (0.4, 11)
                            (0.5, 11)
                            (0.6, 11)
                            (0.7, 11)
                            (0.8, 11)
                            (0.9, 11)
                            (1.0, 11)
                        };
                        \addplot[magenta, dashed] coordinates{
                            (0.0, 12)
                            (0.2, 12)
                            (0.3, 12)
                            (0.4, 12)
                            (0.5, 12)
                            (0.6, 12)
                            (0.7, 12)
                            (0.8, 12)
                            (0.9, 12)
                            (1.0, 12)
                        };
                        \addplot[blue, mark=x] coordinates{
                            (0.0, 21.0)
                            (0.2, 24.7)
                            (0.3, 25.4)
                            (0.4, 25.7)
                            (0.5, 27.2)
                            (0.6, 24.9)
                            (0.7, 18.0)
                            (0.8, 17.0)
                            (0.9, 16.6)
                            (1.0, 16.4)
                        };
                        \legend{TopicRank, KEA++, TopicCoRank$_\textit{assign}$};
                    \end{axis}
                \end{tikzpicture}
            }
            \subfigure[Information Science]{
                \begin{tikzpicture}[scale=.7]
                    \pgfkeys{/pgf/number format/.cd, fixed}
                    \begin{axis}[x=0.304152\linewidth,
                                 xtick={0, 0.2, 0.4, ..., 1},
                                 xmin=0,
                                 xmax=1,
                                 xlabel=$\lambda_k$,
                                 x label style={yshift=.34em},
                                 y=0.0050692\linewidth,
                                 ytick={0, 10, 20, ..., 100},
                                 ymin=0,
                                 ymax=60,
                                 ylabel=f1-score (\%),
                                 y label style={yshift=-.5em}]
                        \addplot[cyan] coordinates{
                            (0.0, 12)
                            (0.2, 12)
                            (0.3, 12)
                            (0.4, 12)
                            (0.5, 12)
                            (0.6, 12)
                            (0.7, 12)
                            (0.8, 12)
                            (0.9, 12)
                            (1.0, 12)
                        };
                        \addplot[magenta, dashed] coordinates{
                            (0.0, 9)
                            (0.2, 9)
                            (0.3, 9)
                            (0.4, 9)
                            (0.5, 9)
                            (0.6, 9)
                            (0.7, 9)
                            (0.8, 9)
                            (0.9, 9)
                            (1.0, 9)
                        };
                        \addplot[blue, mark=x] coordinates{
                            (0.0, 17.6)
                            (0.1, 18.4)
                            (0.2, 18.3)
                            (0.3, 18.6)
                            (0.4, 19.2)
                            (0.5, 19.5)
                            (0.6, 19.4)
                            (0.7, 11.8)
                            (0.8, 8.7)
                            (0.9, 7.7)
                            (1.0, 7.5)
                        };
                    \end{axis}
                \end{tikzpicture}
            }
            \subfigure[Archaeology]{
                \begin{tikzpicture}[scale=.7]
                    \pgfkeys{/pgf/number format/.cd, fixed}
                    \begin{axis}[x=0.304152\linewidth,
                                 xtick={0, 0.2, 0.4, ..., 1},
                                 xmin=0,
                                 xmax=1,
                                 xlabel=$\lambda_k$,
                                 x label style={yshift=.34em},
                                 y=0.0050692\linewidth,
                                 ytick={0, 10, 20, ..., 100},
                                 ymin=0,
                                 ymax=60,
                                 ylabel=f1-score (\%),
                                 y label style={yshift=-.5em},
                                 legend style={font=\footnotesize}]
                        \addplot[cyan] coordinates{
                            (0.0, 21)
                            (0.2, 21)
                            (0.3, 21)
                            (0.4, 21)
                            (0.5, 21)
                            (0.6, 21)
                            (0.7, 21)
                            (0.8, 21)
                            (0.9, 21)
                            (1.0, 21)
                        };
                        \addplot[magenta, dashed] coordinates{
                            (0.0, 18)
                            (0.2, 18)
                            (0.3, 18)
                            (0.4, 18)
                            (0.5, 18)
                            (0.6, 18)
                            (0.7, 18)
                            (0.8, 18)
                            (0.9, 18)
                            (1.0, 18)
                        };
                        \addplot[blue, mark=x] coordinates{
                            (0.0, 31.4)
                            (0.1, 35.1)
                            (0.2, 35.5)
                            (0.3, 35.6)
                            (0.4, 36.7)
                            (0.5, 39.0)
                            (0.6, 33.0)
                            (0.7, 23.5)
                            (0.8, 21.4)
                            (0.9, 20.4)
                            (1.0, 20.4)
                        };
                    \end{axis}
                \end{tikzpicture}
            }
            \caption{
                Behavior of TopicCoRank$_\textit{assign}$ depending on $\lambda_k$ ($\lambda_t = 0.1$)
                \label{fig:lambda_k_variation}
            }
        \end{figure}
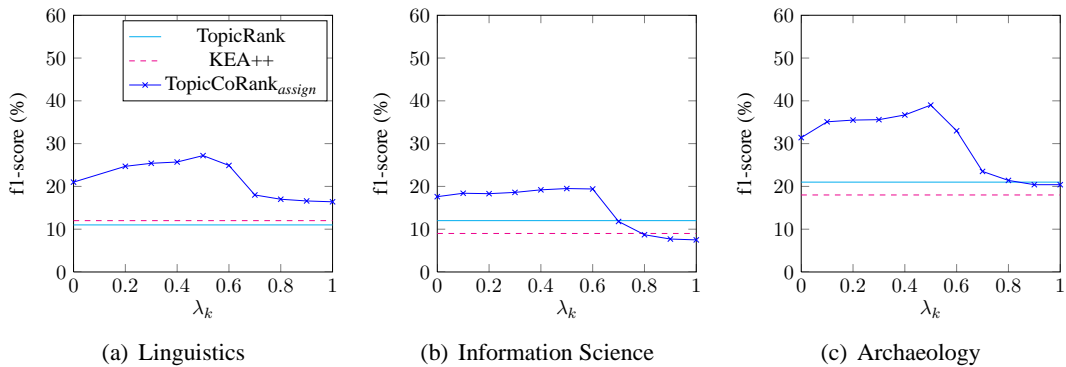

        Fig.~\ref{fig:lambda_k_variation} shows the behavior of TopicCo\-Rank$_\textit{assign}$ when $\lambda_k$ varies from 0 to 1.
        When $\lambda_k = 0$, only the document influences the ranking of the controlled keyphrases.
        As for TopicCoRank$_\textit{extr}$ when $\lambda_t = 0$, TopicCoRank$_\textit{assign}$ is slightly similar to KEA++ when $\lambda_k = 0$.
        When $\lambda_k = 1$, TopicCoRank$_\textit{assign}$ always outputs the same keyphrases: the ones that are the most important in the domain.
        The first half of the curve increases, showing that the relations between the controlled keyphrases have a positive influence on the ranking of the controlled keyphrases.
        Conversely, the second half of the curve decreases.
        Thus, the sole domain is not sufficient for keyphrase annotation.
        
  \subsection{Qualitative example}
\label{subsec:qualitative_example}
    \REVIEWOK{We need the authors to explain clearly and deeply why TopicCoRank can outperform TopicRank}
    To show the benefit of TopicCoRank, we compare it to TopicRank on one of our bibliographic records in Linguistics (see Figure~\ref{fig:example}).
    Over the nine reference keyphrases, TopicRank successfully identifies two of the reference keyphrases: ``lexical semantics'' and ``semantic variation''. TopicCoRank successfully identifies seven of them: ``lexical semantics'', ``verb'', ``semantic variation'', ``French'', ``syntax'', ``semantic interpretation'' and ``distributional analysis''.
    
    \begin{figure}
      \centering
      \framebox[\linewidth]{
        \parbox{.99\linewidth}{\footnotesize
          \textbf{\textit{Toucher : le tango des sens. Problèmes de sémantique lexicale} (The French verb 'toucher': the tango of senses. A problem of lexical)}\\
    
          \textit{A partir d'une hypothèse sur la sémantique de l'unité lexicale 'toucher' formulée en termes de forme schématique, cette étude vise à rendre compte de la variation sémantique manifestée par les emplois de ce verbe dans la construction transitive directe 'C0 toucher C1'. Notre étude cherche donc à articuler variation sémantique et invariance fonctionnelle. Cet article concerne essentiellement le mode de variation co-textuelle : en conséquence, elle ne constitue qu'une première étape dans la compréhension de la construction des valeurs référentielles que permet 'toucher'. Une étude minutieuse de nombreux exemples nous a permis de dégager des constantes impératives sous la forme des 4 notions suivantes : sous-détermination sémantique, contact, anormalité, et contingence. Nous avons tenté de montrer comment ces notions interprétatives sont directement dérivables de la forme schématique proposée.}\\
    
          \textbf{Keyphrases~}:
            \textit{Français} (French); \textit{modélisation} (modelling); \textit{analyse distributionnelle} (distributional analysis); \textit{interprétation sémantique} (semantic interpretation); \textit{variation sémantique} (semantic variation); \textit{transitif} (transitive); \textit{verbe} (verb); \textit{syntaxe} (syntax) and \textit{sémantique lexicale} (lexical semantics).
        }
      }
      \caption{Example of a bibliographic record in Linguistics (\url{http://cat.inist.fr/?aModele=afficheN&cpsidt=16471543})\label{fig:example}}
    \end{figure}

    TopicCoRank mostly outperforms TopicRank because it finds key\-phrases that do not occur within the document: ``French'', ``syntax'', ``semantic interpretation'', and ``distributional analysis''.
    Some keyphrases, such as ``French'', are frequently assigned because they are part of most of the bibliographic records of our dataset\footnote{Yet, TopicCoRank does not assign ``French'' to every bibliographic records.} (48.9\% of the Linguistics records contain ``French'' as a keyphrase);
    Other keyphrases, such as ``semantic interpretation'', are assigned thanks to their strong connection with controlled keyphrases occurring in the abstract (e.g.~``lexical semantics'').

    Interestingly, the performance of TopicCoRank is not only better thanks to the assignment.
    For instance, we observe keyphrases, such as ``verb'', that emerge from topics connected to other topics that distribute importance from controlled keyphrases (e.g.~``semantic variation'').

  \section{Conclusion}
\label{sec:conclusion}
     In this paper, we have proposed a co-ranking approach to performing keyphrase
  extraction and keyphrase assignment jointly. Our method,
  TopicCoRank, builds two graphs: one with the
  document topics and one with controlled keyphrases (training keyphrases). We
  designed a strategy to unify the two graphs and rank by importance topics and
  controlled keyphrases using a co-ranking vote.
  We performed experiments on three datasets of different domains.
  Results showed that our approach benefits from both controlled
  keyphrases and document topics, improving both keyphrase extraction and keyphrase assignment
  baselines. TopicCoRank can be used to annotate keyphrases in scientific domains in a close way of professional indexers.
    
    %

  \section*{Acknowledgments}
  This work was supported by the French National Research Agency (TermITH project – ANR-12-CORD-0029) and by the TALIAS project (grant of CNRS PEPS INS2I 2016, \url{https://boudinfl.github.io/talias/}).

  \bibliography{biblio}
  \bibliographystyle{coling2016}
\end{document}